\documentclass[runningheads]{llncs}
\usepackage[T1]{fontenc}
\usepackage{graphicx,expex,subcaption, booktabs, amsmath}
\usepackage[breaklinks=true]{hyperref}
\usepackage{breakcites}
\lingset{aboveglftskip=-.2ex,interpartskip=\baselineskip,everyglb=\footnotesize}

\begin{document}
\title{Evaluating the Usage of African-American Vernacular English in Large Language Models}
\titlerunning{Evaluating Usage of AAVE in LLMs}
\author{Deja Dunlap \and
R. Thomas McCoy}

\authorrunning{D. Dunlap, R.T. McCoy}
\institute{Yale University, New Haven CT 06511, USA}
\maketitle              % typeset the header of the contribution

\begin{figure}
  \includegraphics[width=\textwidth]{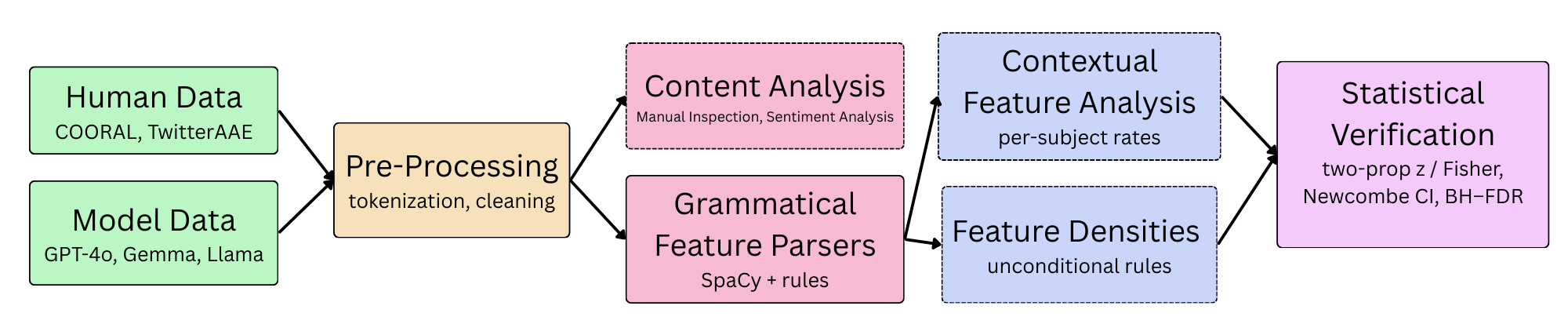}
  \caption{The workflow we used to compare how African-American Vernacular English is used by humans and large language models. This workflow is expanded in further detail in the \textbf{Methodology} section below.}
  \label{fig:method_overview}
\end{figure}

\begin{abstract}
In AI, most evaluations of natural language understanding tasks are conducted in standardized dialects such as Standard American English (SAE). In this work, we investigate how accurately large language models (LLMs) represent African American Vernacular English (AAVE). We analyze three LLMs to compare their usage of AAVE to the usage of humans who natively speak AAVE. We first analyzed interviews from the Corpus of Regional African American Language and TwitterAAE to identify the typical contexts where people use AAVE grammatical features such as \textit{ain’t}. We then prompted the LLMs to produce text in AAVE and compared the model-generated text to human usage patterns. We find that, in many cases, there are substantial differences between AAVE usage in LLMs and humans: LLMs usually underuse and misuse grammatical features characteristic of AAVE. Furthermore, through sentiment analysis and manual inspection, we found that the models replicated stereotypes about African Americans. These results highlight the need for more diversity in training data and the incorporation of fairness methods to mitigate the perpetuation of stereotypes.\footnote{The code for this project, as well as further numerical data, can be found at this link: \url{https://github.com/dejadunlap/AAVE_AI_Human_Usage}}

\end{abstract}

\section{Introduction}
Recent advances in Natural Language Processing (NLP) have led to impressive improvements across various Natural Language Understanding (NLU) tasks, including text summarization, question answering, and sentiment analysis
\cite{devlin2019bert,Radford2019LanguageMA,openai2023gpt4,touvron2023llama,anthropic2024claude}. However, the rapid progress of large language models (LLMs) has also intensified concerns about the propagation of biases embedded within these systems \cite{blodgett2020language,salinas2024s}. The aspect of bias that we focus on is racial linguistic bias, where AI systems are less adept at processing minoritized dialects in comparison to majority dialects \cite{koenecke2020racial,zhou2025disparities}.

In this study, we systematically evaluate the ability of three LLMs---Llama-3-8B-Instruct-Turbo, GPT-4o-mini, and Gemma-2-27B-IT---to replicate the grammatical structures characteristic of African-American Vernacular English (AAVE). We investigated AAVE due to its widespread use within African American communities and its well-documented marginalization in both computational and linguistic research \cite{blodgett2020language,rickford1999aave}. 

We evaluated seven linguistic features of AAVE and compared human usage of these features to model usage to evaluate whether LLMs accurately capture the usage of these features. Importantly, our framework does not require human annotators, reducing potential biases associated with the annotators' regional or sociocultural backgrounds %of the annotators 
and enabling more scalable evaluation. To our knowledge, this is the first scalable, replicable framework for systematically evaluating LLMs on dialectal representation. This also serves as the first quantitative comparison of LLM AAVE usage against human corpora across multiple model families. Additionally, our framework could be adapted to other minoritized dialects. Through this novel evaluative paradigm, we aim to improve the field's understanding of current LLM AAVE usage, allowing the field to measure progress in developing models that accurately capture linguistic diversity, ultimately fostering greater trust, engagement, and inclusivity across user demographics.

We find empirical evidence of both linguistic misrepresentation and stereotype reproduction in model outputs. Our findings suggest that LLMs generally misrepresent the frequency with which these features appear as well as the contexts in which these features are commonly used. For instance, the Llama model underrepresented the usage of the features negative concord, multiple modals, perfective \textit{done}, habitual \textit{be}, double comparatives, and null copula when compared to human usage of the same features, sometimes by large margins (e.g., Llama using habitual \textit{be} only 14\% as often as humans do). Our findings suggest that, while these models seem to capture some version of AAVE, their usage of AAVE does not align with how native AAVE speakers use the dialect.

\section{Related Work}

\subsection{Societal Biases in AI Systems}

Statistical learning systems, such as neural networks, are trained to internalize the statistical regularities that appear in their training data. This trait accounts for much of the power of statistical learning, since it allows a system to encode information about the structure of its target domain. However, one undesirable consequence of it is that such systems may capture the societal biases and prejudices that are prevalent in many naturalistic datasets \cite{gallegos2024bias}. For instance, they may encode negative attitudes about groups that are marginalized based on race \cite{nadeem2021stereoset,an2025measuring}, gender identity \cite{bolukbasi2016man,caliskan2017semantics}, sexual orientation \cite{felkner-etal-2023-winoqueer}, or other factors \cite{nangia2020crows}.
Machine learning systems may also perform less well for users who belong to marginalized groups, likely due to these groups being underrepresented in the training data \cite{buolamwini2018gender,blodgett2017racial}.
Many approaches have been shown to reduce the extent of societal biases in AI systems \cite{lu2020gender,garimella2021he,mattern2022understanding}, but fully removing them is a challenging problem \cite{gonen2019lipstick}. 

Our work's contribution to this line of literature lies in analyzing how LLMs replicate the grammatical properties of a marginalized dialect. This direction relates to bias because one way in which AI systems could create algorithmic injustice is by failing to encode marginalized dialects accurately, resulting in text that presents an inaccurate or caricatured version of the dialect. With the increase of usage of LLMs in content generation \cite{vanEss2025DetectingAIGeneratedContent,Siewert2025AISlop}, this could have dire consequences in the mainstream perception of AAVE and its speakers.

\subsection{AAVE in LLMs}

Within the broad topic of bias in AI systems, our work focuses specifically on the representation of a non-standard dialect, AAVE. Prior studies suggest that LLMs tend to default to Standard American English (SAE) and that attempts to generate non-standard varieties such as AAVE often result in grammatical inaccuracies and the reinforcement of harmful stereotypes \cite{Fleisig:24,lin2024one,aavenue}. 
Further, AI systems tend to perform worse on NLU tasks with AAVE prompts than with SAE prompts \cite{lin2024one,aavenue}.  Our work differs from these prior papers in that we consider the properties of AAVE as its own linguistic tradition rather than its relationship with SAE. Toward this end, we evaluate how computational systems use AAVE in relation to human usage of the dialect. 

\subsection{Comparing LLM Language to Human Language}

An extensive body of literature has investigated the extent to which LLMs capture the properties of human language. Much of this literature evaluates LLMs on whether they have learned the rules that govern syntactic phenomena \cite{marvin2018targeted,warstadt2020blimp} such as subject-verb agreement \cite{linzen-etal-2016-assessing,gulordava2018colorless} and filler-gap dependencies \cite{wilcox-etal-2018-rnn,wilcox2024using}. Our work differs from such papers because we analyze how frequently grammatical phenomena are used and in which contexts, rather than analyzing adherence to the rules of grammar. While  \cite{sandler2024linguistic} and \cite{munoz2024contrasting} also work to compare human spoken text to LLM-generated text and compare the statistical properties of LLM text to the statistical properties of human text, our work differs from these papers in that we conduct such analyses for AAVE.

The most closely related prior work is \cite{Fleisig:24}, which also includes analyses of the dialectal features used by an LLM when producing text in that dialect. \cite{Fleisig:24} analyzes many different dialects, including AAVE, finding that when asked questions by native speakers of various regional and cultural dialects, the models defaulted to responses in SAE. Our work differs from \cite{Fleisig:24} in that we focus specifically on AAVE instead of a range of multiple dialects, enabling us to go into greater depth by providing fine-grained analyses of the features that we identify. In addition, we also extend this prior work by analyzing LLMs from multiple families (Llama, Gemma, and GPT) rather than the single GPT family investigated by \cite{Fleisig:24}.

\section{Methodology}

\subsection{Human-Generated Language}

We utilized the Corpus of Regional African American Language (CORAAL) \cite{coraal2023} and TwitterAAE  \cite{Blodgett2018UDTwitterAAE} datasets for our human baseline datasets. The CORAAL data consists of 272 transcribed sociolinguistic interviews with native AAVE speakers born between the years of 1890 and 2005 from a variety of Eastern and Southern cities. While LLMs are often presumed to be largely trained on written text, the increased usage of LLMs in content generation--such as in the generation of scripts--warrants discussion of LLMs' ability to replicate natural speech. TwitterAAE consists of 1.1 million tweets mostly generated in the United States in 2013 from users who were deemed likely to be using AAVE-associated terms. Due to limitations in budget, we randomly selected 110,000 tweets from the original dataset and treated this as our ``TwitterAAE'' dataset. 

Both datasets were filtered to exclude non-alphanumeric characters. When a character could not be processed by our scripts, the word containing the character was ignored.
After data preprocessing was complete, the CORAAL dataset contained 109,234 sentences and 6,535,668 characters from 272 unique interviews. The TwitterAAE dataset contained 133,288 sentences and 4,605,484 characters from 111,044 unique tweets.

\begin{table}[ht!]
\centering
\caption{
    Prompt templates used to elicit AAVE data in two distinct contexts: extended sociolinguistic interviews (CORAAL) and informal social media posts (TwitterAAE). The variables \textit{gender} and \textit{city} were sampled at a rate proportional to the representation in the CORAAL dataset to control for gender and geographical-related variance.
  }
  \label{tab:prompt_templates}
  \begingroup
  \setlength{\tabcolsep}{10pt}
  \begin{tabular}{|p{0.40\textwidth}|p{0.40\textwidth}|}
  \hline
    \textbf{CORAAL-style Prompt} & \textbf{TwitterAAE-style Prompt} \\
    \hline
    \textbf{Role:} You are an African American \{gender\} from \{city\}, participating in an oral sociolinguistic interview.

    \textbf{Guidelines:}
    \begin{itemize}
        \setlength\itemsep{0.3em}
        \item Speak in a natural, conversational way, as though you are telling your life story, sharing experiences, and reflecting on everyday life.
        \item Produce a continuous narrative roughly the length of a 30-minute conversation (around 4,000–5,000 words).
        \item  Include only the narrative text itself, with no headings, notes, or explanations.
    \end{itemize}
    &
    \textbf{Role:} You are given the role of a casual Twitter user. Generate 250 tweets written in African American Vernacular English (AAVE).

    \textbf{Guidelines:}
    \begin{itemize}
        \setlength\itemsep{0.3em}
        \item Write them in the informal, conversational style of Twitter.
        \item Keep each tweet short (under 280 characters).
        \item Use natural, everyday topics (music, sports, friends, emotions, funny observations, etc.).
        \item Put each tweet on a new line using the newline character.
        \item Include no other extraneous information about the tweet, the task, or anything else. Include only the tweets.
    \end{itemize} \\
    \hline
  \end{tabular}
  \endgroup
\end{table}

\subsection{Model-Generated Text}

The model-generated datasets were created utilizing the prompts listed in \textbf{Table~\ref{tab:prompt_templates}}. Though LLMs can be sensitive to the wording of prompts, we do not expect the details of our prompts to substantially influence our results because the prompts do not refer to specific grammatical features. For the CORAAL-like datasets, we generated 1,000 unique interviews. Similarly, for the TwitterAAE-like datasets, we generated 750 sets of 250 unique tweets ($\approx$187,500 tweets in total). All datasets were filtered similarly to the human datasets. We chose to create more content in the model datasets than was present in the human datasets to create a representative picture of the models' handling of the grammatical features of AAVE. Although the CORAAL prompt is expected to generate a more narrative-style text than the human dataset (by the stop-and-start nature of an interview), this is unlikely to affect the usage of the grammatical features we are observing; rather, it is mainly likely to reflect a difference in the flow of content.  

We applied this process to three LLMs: GPT-4o-mini \cite{openai2023gpt4} accessed via the OpenAI API, and Llama-3-8B-Instruct-Turbo \cite{touvron2023llama} and Gemma-2-27B-IT \cite{team2024gemma} accessed via the Together API. We chose these models because of their widespread popularity and open-access nature of the models.

After data pre-processing, the CORAAL-like Llama 3 corpus contained 53,329 sentences and 3,884,340 characters; the CORAAL-like GPT-4o-mini corpus contained 97,874 sentences and 9,041,315 word tokens; and the CORAAL-like Gemma 2 corpus contained  59,590 sentences and 4,038,365 characters. The TwitterAAE-like Llama 3 corpus contained 39,317 sentences and 1,653,266 characters; the TwitterAAE-like GPT-4o-mini corpus contained 177,797 sentences and 6,355,059 characters; and the TwitterAAE-like Gemma 2 corpus contained 66,396 sentences and 1,695,964 characters.

\section{Experiment 1: Content Analysis}

\subsection{Qualitative analysis of themes}

To provide a qualitative sense of the themes present in model-generated text, we randomly selected 10 interviews in each of the CORAAL-like datasets and 250 tweets in each of the TwitterAAE-like datasets for manual evaluation. While the shorter form TwitterAAE-like data was generally unproblematic with respect to topic, the CORAAL-like data revealed some concerning trends. The models often described stereotypical themes of coming from single-parent homes (\textbf{Example~\ref{ex:single_parent}}), being raised on the ``wrong'' side of town (\textbf{Example \ref{ex:wrong_side}}), and having to do ``anything it took'' to survive the ``streets''.  Across all of the CORAAL-like datasets, there were often descriptions of economic struggle (\textbf{Example \ref{ex:economic_struggle}}) and community violence (\textbf{Example  \ref{ex:violence}}). 

\ex\label{ex:single_parent}\textbf{Gemma:} My mama, she was a pillar, worked her tail off at the post office. Never complained, always made sure we had a hot meal on the table and a roof over our heads. My daddy, well, he wasn't always around.
\xe
\ex\textbf{Llama:} The drug dealers on the corners, the sirens wailing in the night, the whispers of violence hanging in the air. It’s like the city has this shadow side, always lurking beneath the surface. \label{ex:wrong_side} 
\xe

\ex\textbf{Gemma:} Growing up, it was tough, you know? We didn't have much, but we had each other.
\label{ex:economic_struggle}
\xe

\ex \label{ex:violence}\textbf{Llama:} Our neighborhood was... tough. Lots of poverty, gangs, crime.
\xe
Furthermore, even when adopting a more positive tone, the models seemed to default to stereotypical notions of the need to ``escape'' from the ``hood'' through basketball (\textbf{Example \ref{ex:basketball}}) or the military (\textbf{Example \ref{ex:military}}).

\ex \label{ex:basketball}\textbf{GPT-4o:} I played basketball though—man, I loved that sport. It was an escape.
\xe

\ex \label{ex:military}\textbf{GPT-4o:} I decided to join the military. Man, that was a big decision...but I knew it was the right choice.
\xe

Conversely, the models occasionally reflected a deep appreciation for Black history and culture. The models even sometimes mentioned a deep pride in the use of AAVE (\textbf{Example \ref{ex:AAVE_pride}}), expressing a need to continue a rich linguistic tradition for future generations. Many model interviews also stressed the importance of giving back to one's community through activism and/or mentorship (\textbf{Example \ref{ex:mentoring}}).

\ex\label{ex:AAVE_pride}\textbf{Llama:} I'm proud of our history, our traditions, our people. We might not be perfect, but we're imperfectly perfect, you know what I'm sayin'? \xe
\ex\label{ex:mentoring}\textbf{GPT-4o:} Over time, I began to feel a strong pull towards giving back. I wanted to take all that I had learned, all that had been gifted to me, and share it with younger generations. I got involved in mentoring programs for young Black girls, helping them see their potential. \xe
Within each model-generated dataset, we observed substantial similarity across the stories, leading to many repeated phrases across the texts.

\subsection{Quantitative analysis of sentiment} 
\begin{figure}[h!]
  \centering
  \includegraphics[width=0.90\linewidth,alt={Bar charts showing sentiment by corpus; humans show more negative and neutral, while models—especially GPT-4o—skew positive.}]{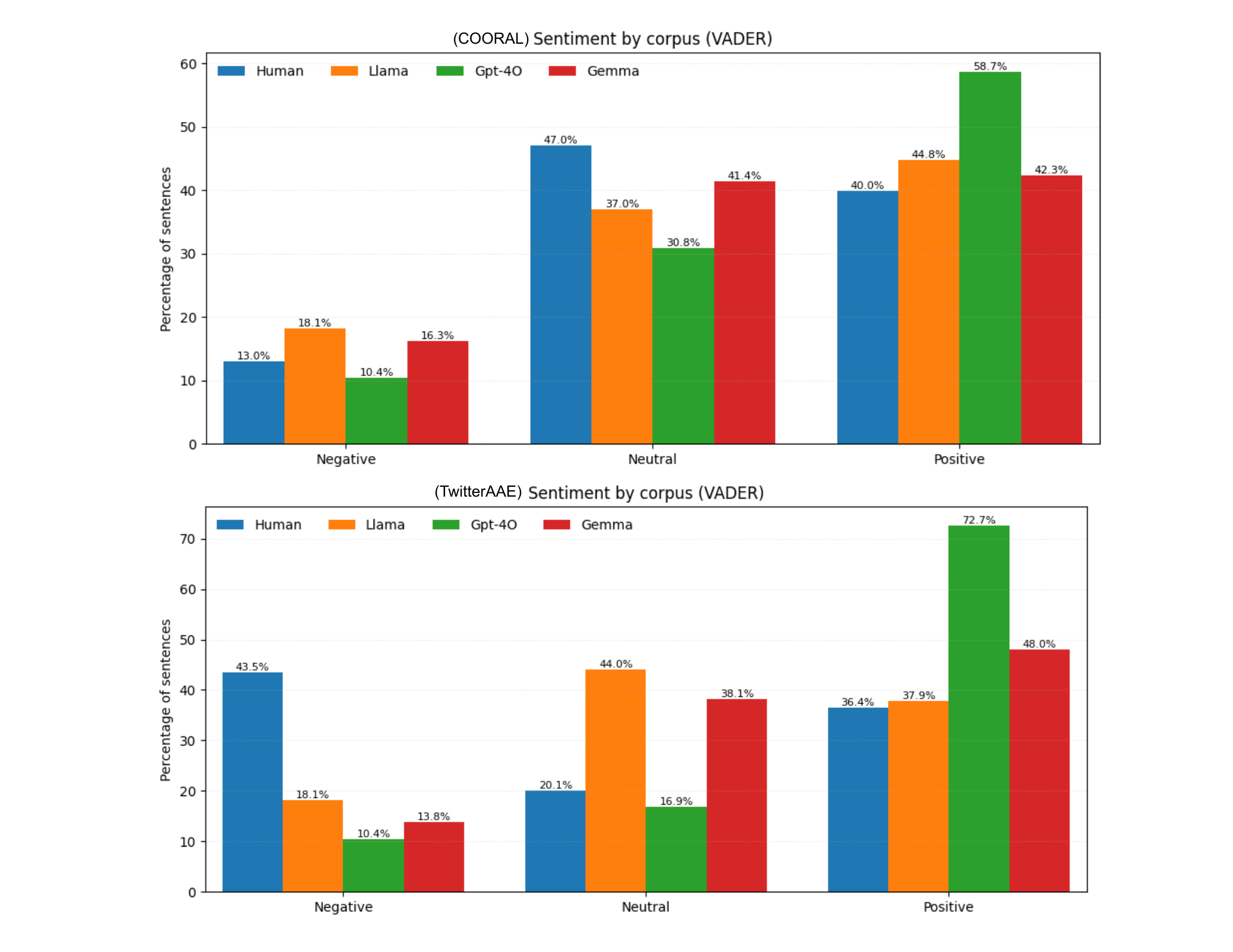}
  \caption{Bar plots highlighting the portion of sentences in each corpus that had a sentiment score $x$ defined as negative ($x<-0.5$), neutral ($0.5> x >-0.5$), or positive ($x>0.5$), where $x$ was obtained via VADER.}
  \label{fig:plot_sentiment}
\end{figure}

We utilized sentiment analysis to summarize the general emotion associated with a piece of text, with VADER (Valence Aware Dictionary and sEntiment Reasoner) \cite{hutto2014vader}. As shown in \textbf{Table~\ref{tab:sentiments}}, the average sentence-level sentiments across the datasets were generally in the same range, with GPT-4o having the highest (and thus most positive) mean score. One point of note is that, as reflected in  \textbf{Figure~\ref{fig:plot_sentiment}}, the models generally had a higher proportion of sentiments that fell in the `positive' category than the human dataset, with GPT-4o having almost 20\% more positive sentences across the dataset than the human dataset. Consequently, the proportion of `neutral' and `negative' sentences was lower in the model datasets than in the human dataset. Overall, these results show that the LLMs and humans were fairly similar in what the overall `positivity' in their texts tended to be, except for GPT.

\begin{table}[ht!]
\caption{Descriptive data for sentence-level sentiment scores across datasets. Scores range from -1 (furthest negative) to +1 (furthest positive).}
  \label{tab:sentiments}
  \centering
  \begin{tabular}{ | p{6em} | p{6em} | p{6em} | p{6em} | p{6em} | }
  \hline
        \textbf{Sentiment} & Human & Gemma & GPT-4o & Llama  \\
        \hline
        Average       & 0.135 & 0.146 & 0.260 & 0.160\\
    \hline
  \end{tabular}
\end{table}
\section{Experiment 2: Feature Frequency}

We now turn to analysis of several grammatical features that are characteristic of AAVE, investigating whether LLMs and humans use these features with the same frequencies and in the same ways.

\subsection{Linguistic Phenomena}

This paper focuses on seven grammatical features of AAVE, illustrated in \textbf{Table~\ref{tab:grammatical_features}}: the negation word \textit{ain't}, negative concord, habitual \textit{be}, double comparatives, perfective \textit{done}, multiple modals, and null copula \cite{green2002african,YGDPProject2025}. While these features are commonly associated with informal or nonstandard language,  they serve specific semantic and syntactic roles in AAVE \cite{green2002african}. We chose these features for several reasons. First, they are widely used in AAVE and other dialects but are not present in Standard American English (SAE). Second, even though it is well-established that they are just as legitimate and systematic as grammatical phenomena in SAE, these features have historically been stigmatized as improper or ungrammatical, contributing to negative stereotypes about communities that use them prominently. Additionally, these features occurred frequently enough in the human corpus that we could estimate their frequency with reasonable confidence, avoiding sparsity as a pitfall in our analysis. Finally, these features have documented usage as far back as the 1950s \cite{green2002african,YGDPProject2025}, but are thought to have been used long before then. In this way, we are able to ensure that the features we are observing will be reflected across both of the datasets, regardless of the temporal and geographical differences.

\begin{table}
\caption{
    Examples of the grammatical features of African-American Vernacular English (AAVE) that are examined in this study, along with analogous sentences in Standard American English (SAE).
  }\label{tab:grammatical_features}
  \noindent
  \centering
  \resizebox{\textwidth}{!}{
  \setlength\tabcolsep{0.2cm}
  \begin{tabular}{lp{5.2cm}p{5cm}}
  \toprule
    \textbf{AAVE Feature}           & \textbf{Example Usage of Feature} & \textbf{Analogous Sentence in SAE} \\
    \midrule
    \textit{Ain't}       & I \textbf{ain't} doing all that.         &     I \textbf{am not} doing all of that.                    \\
    \midrule
    Habitual \textit{Be}     & He \textbf{be} doing too much.       & He \textbf{is always} doing too much.                \\
    \midrule
    Negative Concord & I \textbf{don't never} have \textbf{no} problems.          &  I \textbf{don't} ever have any problems.                     \\     \midrule

    Double Comparative       & I am \textbf{more happier} when I'm alone.         &    I am \textbf{happier} when I'm alone.                  \\     \midrule

    Perfective \textit{Done}       & I \textbf{done} lost my wallet.         &     I \textbf{have} lost my wallet.                 \\
        \midrule

    Multiple Modals      & We \textbf{might can} go up there next Saturday.         &   We \textbf{might be able} to go up there next Saturday.       \\
        \midrule

    Null Copula      & Your mama \textbf{$\emptyset$} a weight-lifter.         &   Your mama \textbf{is} a weight-lifter.      \\
    \bottomrule
  \end{tabular}}
\end{table}

\subsection{Grammatical Feature Detection}
We determine the frequency of the AAVE grammatical features in across human and model-generated texts. To determine whether a given feature appeared in a text, we developed feature-detection algorithms based on the grammatical features' properties as outlined in \cite{YGDPProject2025}. These algorithms incorporated grammatical features extracted with spaCy (\url{https://spacy.io/}). \textbf{Table \ref{tab:feature_performance}} details the accuracy of the feature detectors on a small (ex. 500 sentences) hand-annotated dataset from CORAAL.

\begin{table}[ht!]
\centering
\caption{Performance of feature detectors across grammatical categories. All feature parsers can maintain an accuracy rate greater than 90.}
\label{tab:feature_performance}
\begin{tabular}{|l|p{0.12\textwidth}|p{0.1\textwidth}|p{0.1\textwidth}|p{0.1\textwidth}|p{0.1\textwidth}|p{0.12\textwidth}|p{0.1\textwidth}|}
\hline
 & Double Comparative & Habitual \textit{Be} & Multiple Modals & Negative Concord & Null Copula & Perfective \textit{Done} & \textit{Ain't} \\
\hline
Accuracy & 0.99 & 0.99 & 0.99 & 0.97 & 0.96 & 0.99 & 1.00  \\
\hline
\end{tabular}
\end{table}

\subsection{Statistical Analysis}

To statistically verify our results in human-model gaps in frequencies, we formulated the density of each feature as the proportion of sentences in which the feature was detected. For each feature, we compared model and human densities with a two-proportion test (equivalently, a $\chi^{2}$ test with 1 df). When any expected cell count was $<$ 5, we used Fisher’s exact test instead. We report effect sizes as the difference (model - human) with $95$ Newcombe/Wilson CIs.

\begin{table}[t]
\scriptsize
\caption{Feature frequencies across TwitterAAE and CORAAL (and analogous
datasets generated by LLMs), per 10,000 sentences, rounded to three decimal
points. Bolded $\delta$ columns represent statistically significant differences with $\alpha < 0.05$}\label{ref:feature_frequencies}
\title{Features Densities Across Dataset (per 10,000 sentences)}
\setlength{\tabcolsep}{3pt}
\begin{subtable}[t]{0.35\textwidth}
\caption{CORAAL}
\begin{tabular}{l r r c r c r c}
\toprule
Feature & Human & Gemma & $\Delta$ Gemma & Llama & $\Delta$ Llama & GPT & $\Delta$ GPT \\
\midrule
Habitual \textit{be} 
& 29.661 
& 4.867
& \textbf{-24.795} 
& 13.501
& \textbf{-16.160}
& 11.443
& \textbf{-18.218} \\

Null Copula 
& 354.651 
& 436.650 
& \textbf{81.999} 
& 290.649
& \textbf{-64.003}
& 633.468 
& \textbf{278.816} \\

Perfective Done 
& 11.443 
& 4.363
& \text{-7.080} 
& 1.875
& -9.568 
& 0.102
& \textbf{-11.341} \\

\textit{Ain't} 
& 85.230 
& 253.902 
& \textbf{168.672} 
& 159.388 
& \textbf{74.158}
& 0.715
& \textbf{-84.515} \\

Negative Concord 
& 514.858 
& 188.454 
& \textbf{-326.404}
& 231.394
& \textbf{-283.464}
& 22.580 
& \textbf{-492.278} \\

Double Comparative 
& 3.662 
& 0.168 
& \textbf{-3.494}
& \textbf{0.375} 
& -3.287 
& 0.817
& \textbf{-2.844}\\

Multiple Modals 
& 25.908 
& 0.168
& \textbf{-25.740}
& 0.563
& \textbf{-25.345} 
& 0.204 
& \textbf{-25.703} \\

\bottomrule
\end{tabular}
\end{subtable}

\vspace*{0.25cm}

\begin{subtable}[t]{0.35\textwidth}
\caption{TwitterAAE}
\begin{tabular}{l r r c r c r c}
\toprule
Feature & Human & Gemma & $\Delta$ Gemma & Llama & $\Delta$ Llama & GPT & $\Delta$ GPT \\
\midrule
Habitual \textit{be} 
& 44.115 
& 35.846 
& \textbf{-8.269}
& 4.324 
& \textbf{-39.791}
& 87.853
& \textbf{43.738} \\

Null Copula 
& 254.637 
& 98.349
& \textbf{-156.287}
& 254.343 
& \textbf{-0.294}
& 254.841 
& \textbf{0.205} \\

Perfective Done 
& 13.505 
& 13.404 
& -0.100 
& 66.129 
& \textbf{52.625}
& 3.262
& \textbf{-10.242} \\

\textit{Ain't} 
& 242.332 
& 201.819 
& \textbf{-40.513}
& 880.535 
& \textbf{638.203}
& 116.256 
& \textbf{-126.076} \\

Negative Concord 
& 249.835 
& 71.691 
& \textbf{-178.144} 
& 267.314 
& \textbf{17.479}
& 204.953 
& \textbf{-44.882} \\

Double Comparative 
& 1.125 
& 0.151 
& \textbf{-0.975 }
& 0.254 
& -0.871 
& 0.562 
& -0.563 \\

Multiple Modals 
& 2.176 
& 0.151
& \textbf{-2.025}
& 1.780 
& -0.395 
& 1.069 
& \textbf{-1.107} \\
\bottomrule
\end{tabular}
\end{subtable}
\end{table}

\begin{figure}[!ht]
  \centering
  \includegraphics[width=\textwidth,alt={Bar charts comparing linguistic feature frequencies; humans use AAE features such as 'ain’t' and null copula far more than models.}]{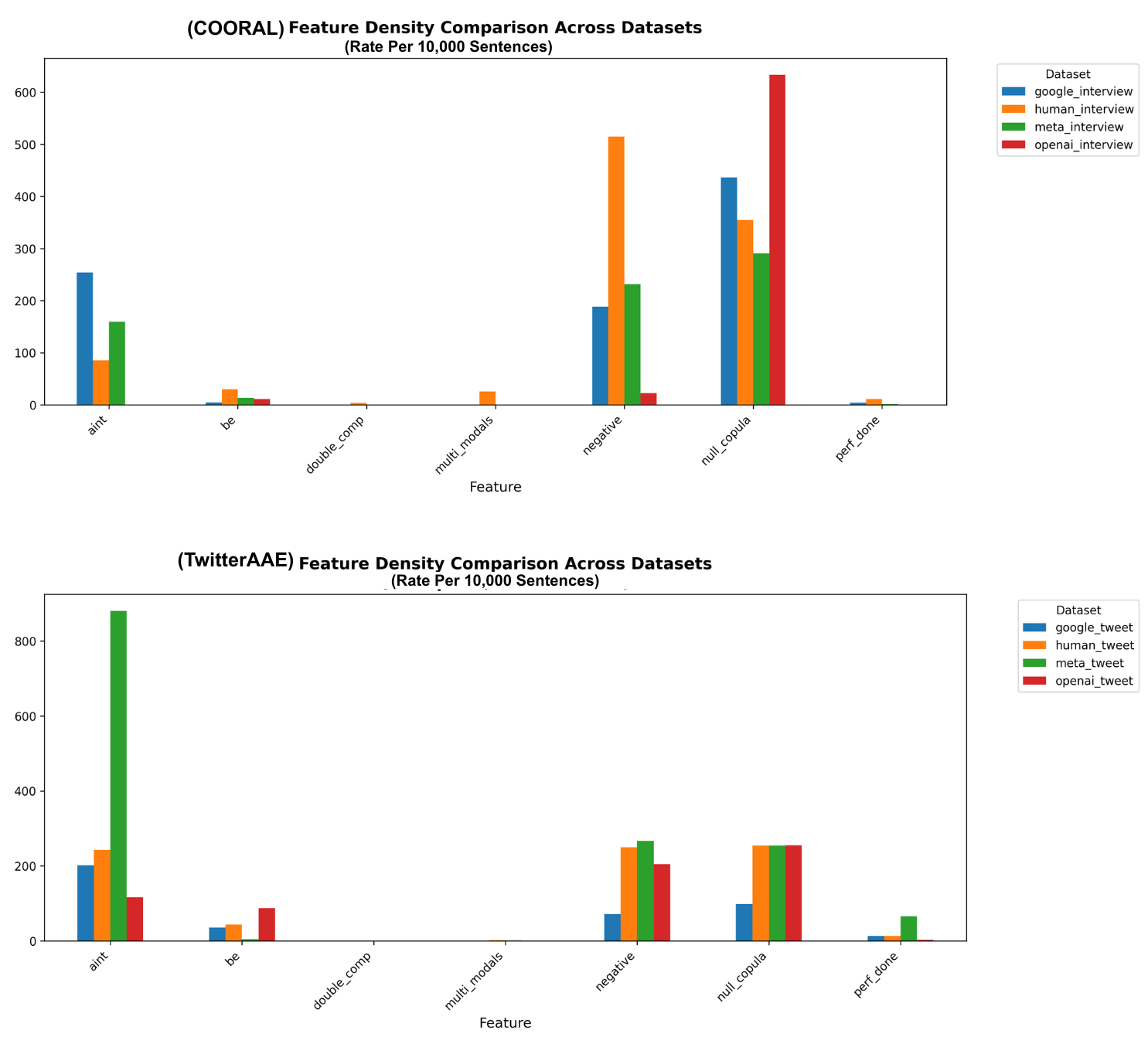}
  \caption{Bar plots highlighting the difference in feature frequency by model and feature (per 10,000 sentences). Features that appear less frequently (ie. Double Comparative, Null Modals, Perfective \textit{done}) have a harder time appearing on scale; see Table~\ref{ref:feature_frequencies} for those features.}
  \label{fig:plot_feature_densities}
\end{figure}

In the CORAAL comparison, across all seven features, model--human gaps are prevalent and are statistically significant in all cases (\textbf{Figure~\ref{fig:plot_feature_densities}}, \textbf{Table~\ref{ref:feature_frequencies}}). Models \emph{underuse} negative concord and habitual \textit{be} relative to humans; the shortfall is largest for GPT-4o and Gemma respectively. Multiple modals, perfective \textit{done}, and double comparatives are also consistently below human rates for all models (though human usage was also relative low). Null copula splits by model: Llama is below human level, while Gemma and GPT-4o are \emph{above} human level (with GPT-4o showing the largest increase). For \textit{ain't}, Llama and Gemma exceed human usage, while GPT-4o is quite far below.

In the TwitterAAE comparison, models are closer to each other but remain meaningfully different from human usage. Negative concord, \textit{ain't}, and perfective \textit{done} have a similar split: they are underused by Gemma and GPT, but overused for Llama. Multiple modals and double comparatives remain uniformly lower than human usage (though still relatively rare). Null copula is \emph{below} human usage for all models (contrasting with CORAAL, where Gemma and GPT-4o were above human usage). Habitual \textit{be} is underused by Gemma and Llama, but overused by GPT.

Models do not modify human usage in a uniform way; they \emph{reweight} AAE features in ways that depend on both the feature and the domain. The largest practical gaps appear for negative concord and \textit{ain't}, while differences in habitual \textit{be}, null copula, and \textit{ain't} reveal model-specific tendencies (e.g., GPT-4o strongly suppresses \textit{ain't}). Even when differences are statistically significant, several differences (e.g., GPT-4o on double comparative in CORAAL) are small in absolute terms, whereas others represent substantial stylistic shifts relative to human baselines.

\section{Experiment 3: Contextual Feature Analysis}

To provide a quantitative analysis of the contexts in which \textit{ain't}, habitual \textit{be}, the perfective \textit{done}, and the null copula are used, we utilized the parser-based feature identification method described in the previous section to determine the subject with which the grammatical feature was used. We did not include negative concord, multiple modals, or double comparatives in this analysis because the usage of these features cannot be localized to a single specific word---rather, they depend on the co-occurrence of multiple related terms. For each grammatical phenomenon, we identified the 10 subjects that are used most frequently by humans with that phenomenon and then analyzed whether models had similar rates of using the phenomenon with that subject.

For the CORAAL comparisons, across all four features, models generally underuse the construction relative to humans \emph{within the same contexts}, with most differences (where possible) being statistically significant (\textbf{Table \ref{tab:contextual_densities_interviews}}). Most strikingly, for many rows across all four features (but especially the perfective \textit{done}), models did not have any usage with the same subjects that humans did. For habitual \textit{be} and null copula, all three models are below human rates for almost every row. The shortfall is largest on frequent pronouns (e.g., \textit{I}, \textit{it}, \textit{she}, \textit{they}), indicating a broad compression of these features across contexts rather than a single misspecified context. The \textit{ain't} feature showed that, relative to human contexts, models suppress usage in most rows. The effect is strongest for first-person and impersonal pronoun contexts (\textit{I} and \textit{it}), suggesting that models avoid \textit{ain’t} precisely where humans most often use it. Perfective \textit{done} shows extremely sparse human usage and near-zero model usage; almost all model–human differences are negative and significant. This indicates a systematic failure to reproduce perfective \textit{done} even in the few contexts where humans use it.

In the TwitterAAE datasets, the same qualitative pattern holds: across all four features, models rarely used the features in the same context that humans did. For context that were shared between human and model usage, the models typically underutilized the grammatical features in the same manner that was most common in human speech.

\begin{table}[h!]
\centering
\caption{Contextual frequencies of grammatical features in the CORAAL datasets (rounded to three decimal points). Values show Human baseline and model deviations. Bold indicates significant deltas. (--) indicates the model did not use the feature in that given context.} \label{tab:contextual_densities_interviews}

\begin{subtable}[t]{0.45\textwidth}
\centering
\caption*{Habitual \textit{Be}}
\resizebox{1.0\columnwidth}{!}{
\begin{tabular}{@{}lcccc@{}}
\toprule
Context & Human & $\Delta$ Gemma & $\Delta$ Llama & $\Delta$ GPT \\
\midrule
everybody & 0.006 & -- & -- & -- \\
he & 0.001 & -- & 0.001 & -- \\
I & 0.002 & \textbf{-0.001} & -0.001 & \textbf{-0.001} \\
it & 0.001 & \textbf{-0.001} & \textbf{-0.001} & \textbf{-0.001} \\
people & 0.001 & -- & -- & -- \\
she & 0.002 & -0.001 & -0.001 & -0.001 \\
that & 0.000 & -- & -- & -- \\
they & 0.003 & \textbf{-0.003} & -0.001 & \textbf{-0.003} \\
we & 0.002 & -0.001 & -0.001 & -0.000 \\
you & 0.001 & -0.001 & -0.000 & 0.000 \\
\bottomrule
\end{tabular}
}
\end{subtable}
\hfill
\begin{subtable}[t]{0.45\textwidth}
\centering
\caption*{Null Copula}
\resizebox{1.0\columnwidth}{!}{
\begin{tabular}{@{}lcccc@{}}
\toprule
Context & Human & $\Delta$ Gemma & $\Delta$ Llama & $\Delta$ GPT \\
\midrule
he & 0.017 & -- & -- & -- \\
him & 0.027 & -- & -- & -- \\
it & 0.012 & \textbf{0.007} & 0.003 & \textbf{0.025} \\
me & 0.027 & -0.002 & -0.004 & \textbf{-0.009} \\
people & 0.026 & 0.003 & 0.005 & 0.006 \\
she & 0.011 & -- & -- & -- \\
that & 0.004 & \textbf{0.012} & -- & \textbf{0.013} \\
they & 0.029 & -- & -- & -- \\
we & 0.020 & \textbf{-0.013} & -- & \textbf{-0.005} \\
you & 0.020 & \textbf{-0.012} & \textbf{-0.015} & 0.001 \\
\bottomrule
\end{tabular}
}
\end{subtable}

\vspace{0.1cm}

\begin{subtable}[t]{0.45\textwidth}
\centering
\caption*{\textit{Ain't}}
\resizebox{1.0\columnwidth}{!}{
\begin{tabular}{@{}lcccc@{}}
\toprule
Context & Human & $\Delta$ Gemma & $\Delta$ Llama & $\Delta$ GPT \\
\midrule
he & 0.004 & -- & -- & -- \\
I & 0.007 & 0.001 & -0.001 & -- \\
it & 0.004 & \textbf{0.028} & \textbf{0.018} & \textbf{-0.004} \\
n*ggas & 0.035 & -- & -- & -- \\
she & 0.004 & -- & -- & -- \\
that & 0.001 & \textbf{0.002} & -- & -- \\
they & 0.005 & 0.003 & 0.000 & -- \\
this & 0.001 & -- & \textbf{0.003} & -- \\
we & 0.004 & \textbf{0.013} & 0.002 & -- \\
you & 0.004 & -0.001 & \textbf{0.002} & -- \\
\bottomrule
\end{tabular}
}
\end{subtable}
\hfill
\begin{subtable}[t]{0.45\textwidth}
\centering
\caption*{Perfective \textit{Done}}
\resizebox{1.0\columnwidth}{!}{
\begin{tabular}{@{}lcccc@{}}
\toprule
Context & Human & $\Delta$ Gemma & $\Delta$ Llama & $\Delta$ GPT \\
\midrule
he & 0.001 & -- & -- & -- \\
I & 0.001 & -- & \textbf{-0.001} & -- \\
it & 0.001 & -0.000 & -- & -- \\
lot & 0.000 & -- & -- & -- \\
name & 0.000 & -- & -- & -- \\
she & 0.001 & -0.000 & -- & -- \\
they & 0.001 & -- & -0.000 & -- \\
we & 0.000 & -- & -- & -- \\
you & 0.000 & -- & -- & -- \\
\bottomrule
\end{tabular}
}
\end{subtable}

\vspace{0.1cm}

\end{table}

For both CORAAL and TwitterAAE, models largely \emph{retain} which contexts are more likely to host a feature (qualitative ordering), but they \emph{compress} the magnitudes, yielding persistent underuse within contexts. The effect is strongest for null copula and perfective \textit{done} (broad suppression across contexts), and clear for habitual \textit{be} and \textit{ain’t}. Across datasets the following pattern holds: contextual distributions look plausible, yet model rates remain significantly lower than human baselines in many rows.

\begin{table}[h!]
\centering
\caption{Contextual frequencies of grammatical features in the TwitterAAE datasets (rounded to three decimal points). Values show Human baseline and model deviations. Bold indicates significant deltas. (--) indicates the model did not use the feature in that given context.}\label{tab:contextual_analysis_tweets}

\begin{subtable}[t]{0.45\textwidth}
\centering
\caption*{Habitual \textit{Be}}
\resizebox{1.0\columnwidth}{!}{
\begin{tabular}{@{}lcccc@{}}
\toprule
Context & Human & $\Delta$ Gemma & $\Delta$ Llama & $\Delta$ GPT \\
\midrule
he & 0.003 & -- & -- & -- \\
hoes & 0.010 & -- & -- & -- \\
I & 0.003 & -0.002 & -0.002 & \textbf{0.002} \\
n*ggas & 0.010 & -- & -- & -- \\
people & 0.007 & -- & -- & 0.011 \\
she & 0.005 & -- & -- & -- \\
they & 0.008 & -- & -- & 0.007 \\
u & 0.000 & -- & -- & -- \\
we & 0.003 & -- & -- & 0.001 \\
you & 0.002 & -- & -- & -- \\
\bottomrule
\end{tabular}
}
\end{subtable}
\hfill
\begin{subtable}[t]{0.45\textwidth}
\centering
\caption*{Null Copula}
\resizebox{1.0\columnwidth}{!}{
\begin{tabular}{@{}lcccc@{}}
\toprule
Context & Human & $\Delta$ Gemma & $\Delta$ Llama & $\Delta$ GPT \\
\midrule
he & 0.031 & -0.006 & -- & -- \\
it & 0.018 & -0.006 & 0.002 & \textbf{0.006} \\
me & 0.016 & \textbf{0.010} & \textbf{0.013} & \textbf{0.013} \\
she & 0.030 & 0.017 & -- & -- \\
that & 0.010 & -0.002 & -- & 0.001 \\
they & 0.033 & -0.017 & -0.013 & -0.002 \\
u & 0.007 & -- & -- & -- \\
we & 0.031 & -- & -0.004 & -0.007 \\
you & 0.026 & -0.009 & -0.009 & -0.002 \\
\bottomrule
\end{tabular}
}
\end{subtable}

\vspace{0.1cm}

\begin{subtable}[t]{0.45\textwidth}
\centering
\caption*{\textit{Ain't}}
\resizebox{1.0\columnwidth}{!}{
\begin{tabular}{@{}lcccc@{}}
\toprule
Context & Human & $\Delta$ Gemma & $\Delta$ Llama & $\Delta$ GPT \\
\midrule
he & 0.021 & -- & -- & \textbf{0.022} \\
I & 0.014 & \textbf{0.018} & \textbf{0.096} & \textbf{-0.009} \\
it & 0.017 & -0.004 & \textbf{0.023} & \textbf{-0.013} \\
n*ggas & 0.027 & -- & -- & -- \\
she & 0.019 & \textbf{0.046} & \textbf{0.097} & -- \\
still & 0.032 & -- & \textbf{0.069} & \textbf{-0.023} \\
that & 0.007 & -- & -- & \textbf{-0.006} \\
u & 0.005 & -- & -- & -- \\
we & 0.019 & -- & \textbf{0.059} & \textbf{-0.010} \\
you & 0.014 & 0.003 & \textbf{0.071} & \textbf{-0.006} \\
\bottomrule
\end{tabular}
}
\end{subtable}
\hfill
\begin{subtable}[t]{0.45\textwidth}
\centering
\caption*{Perfective \textit{Done}}
\resizebox{1.0\columnwidth}{!}{
\begin{tabular}{@{}lcccc@{}}
\toprule
Context & Human & $\Delta$ Gemma & $\Delta$ Llama & $\Delta$ GPT \\
\midrule
all & 0.001 & -- & -- & -- \\
hair & 0.003 & 0.005 & -- & 0.000 \\
he & 0.002 & -- & 0.009 & 0.001 \\
I & 0.001 & -0.000 & \textbf{0.012} & \textbf{-0.001} \\
it & 0.001 & 0.002 & 0.003 & -- \\
she & 0.001 & -- & \textbf{0.021} & -- \\
we & 0.001 & -- & 0.003 & -- \\
who & 0.002 & -- & 0.004 & -- \\
work & 0.002 & -- & -- & -- \\
you & 0.001 & 0.001 & 0.002 & -0.000 \\
\bottomrule
\end{tabular}
}
\end{subtable}

\vspace{0.1cm}

\end{table}

\section{Experiment 4: Controlling for Human Noise}

One potential concern is that the human results might be noisy, such that our estimates of human usage might not be very reliable,  affecting our ability to interpret the models' usage. To test for this possibility, we created 10 mini datasets, where each one is created by randomly selecting 2000 sentences from the full CORAAL dataset, and performed the feature identification procedure described above. As seen in \textbf{Table~\ref{tab:mini_datasets_consistent}}, the frequencies of the grammatical features were similar across datasets. This suggests the features present in the human dataset, while not wholly representative of AAVE, hold stable across the CORAAL dataset regardless of the temporal and geographic variations present within the dataset.

\begin{table}[h]
\centering
\caption{Table showing feature densities (per 100 words) across ten randomly selected mini-datasets generated from the human dataset.}\label{tab:mini_datasets_consistent}
\setlength{\tabcolsep}{3pt}
\scriptsize
\begin{tabular}{lrrrrrrrrrr}
\hline
Feature & V1 & V2 & V3 & V4 & V5 & V6 & V7 & V8 & V9 & V10 \\
\hline
Habitual \textit{be} & 0.576 & 0.441 & 0.401 & 0.493 & 0.525 & 0.413 & 0.483 & 0.408 & 0.488 & 0.359\\
Negative Concord & 2.738 & 2.828 & 2.778 & 2.832 & 2.931 & 2.868 & 2.730 & 2.815 & 2.749 & 2.857\\
\textit{Ain't} & 2.279 & 2.620 & 2.378 & 2.306 & 2.406 & 2.372 & 2.655 & 2.373 & 2.530 & 2.356 \\
Double Comparative & 0.008 & 0.000 & 0.000 & 0.008 & 0.000 & 0.008 & 0.017 & 0.017 & 0.000 & 0.008 \\
Multiple Modals & 0.008 & 0.008 & 0.000 & 0.008 & 0.000 & 0.025 & 0.017 & 0.008 & 0.008 & 0.025 \\
Perfective \textit{Done} & 0.125 & 0.183 & 0.142 & 0.125 & 0.183 & 0.149 & 0.133 & 0.150 & 0.168 & 0.092 \\
Null Copula & 2.763 & 2.878 & 2.795 & 2.807 & 2.881 & 2.744 & 2.929 & 2.715 & 2.698 & 2.699 \\
\hline
\end{tabular}
\end{table}

\section{Discussion}

Previous studies have found that LLMs tend to produce ``stereotyped'' language when prompted to generate text in minoritized dialects, including AAVE \cite{Fleisig:24}. 
While standard post-training methods likely discourage models from producing overtly racist language, our \textbf{Experiment 1} provides evidence that these models retain certain implicit associations between African Americans and stereotypes about their upbringings, congruent with other studies reporting implicit bias embedded in these systems \cite{dacon-etal-2022-evaluating,aavenue,garimella2021he,mattern2022understanding}. 
This phenomenon is likely due to the training data used for these models that reflect discriminatory and stereotypical representations of Black culture \cite{Towns03092015}. However, our sentiment analysis revealed an interesting insight: the sentiment analyses of model-generated datasets yielded more positive sentiment than those of the human dataset. That is to say that while the model-generated data was likely to reflect stereotypical portrayals of African-Americans, it expressed this stereotypical content in an extremely positive manner. LLMs' positivity bias has been noted in the literature \cite{bardol2025chatgptreadstoneresponds}. In the context of this study, it had the effect of making the data lean toward broad melodrama and didactic beats, with archetypal characters and schematic plotting rather than natural human-like dialogues. This positivity bias risks producing melodramatic portrayals of AAVE speakers, which can be equally distancing from authentic usage. These results further support the need for approaches that combat this concerning trend through more targeted training methods \cite{Fleisig:24} or more diversity in training data \cite{Hofmann:24,Ntoutsi}. 

The question of whether large language models (LLMs) can ``understand'' African American Vernacular English (AAVE) is complex. In this study, we evaluated Llama-3-8B, Gemma-2-27B, and GPT-4o-mini on their ability to represent and contextualize seven AAVE features. While there was some general variation in feature presentation across features, contexts, and datasets, one trend was consistent: the models tended to underrepresent grammatical features associated with AAVE.  The fact that the features evaluated had such a varying representation could be because these features have a fairly broad presence across multiple English dialects beyond just AAVE, such as Southern American English \cite{YGDPProject2025}. This overlap across dialects raises the possibility that models are blurring the lines between Black speech and other dialects, leading to diluted representations of AAVE-specific grammar. This possibility would be consistent with previous studies that have shown that LLMs generally have a poorer understanding of AAVE compared to other dialects \cite{dacon-etal-2022-evaluating,aavenue,dacon-2022-towards}. The under-representation of these features may also stem from aspects of post-training that aim to suppress perceived ``improper'', ``nonstandard'', or even ``negative'' language, inadvertently steering the models away from key aspects of AAVE. Paradoxically, this kind of harm-avoidant alignment could reinforce negative linguistic biases by erasing legitimate grammatical patterns. 

Most strikingly, many of the human-specific contextual usages of the features rarely or never appeared in model-generated text. This observation suggests that, while the models might capture these features to some extent, their internalized notion of AAVE speech might not be wholly aligned with how AAVE is used by native AAVE speakers. The reasons for this might also be similar to the general misrepresentation of the features previously described.

Beyond our specific findings, this work highlights the importance of systematically incorporating linguistic and dialectal diversity into AI development. Such measures are important in improving the fairness and representativeness of AI systems.

\subsection{Limitations}

The CORAAL dataset exclusively contained speakers of AAVE along the East Coast, particularly areas with higher dialectal densities. Thus, our findings should be interpreted as illustrative rather than exhaustive, reflecting the features present in these specific corpora rather than AAVE in its entirety. The TwitterAAE dataset, while larger in size, only specified whether a certain text contained elements of AAVE, rather than coming from a native AAVE speaker. This could affect the representation of AAVE present in the TwitterAAE dataset.

Additionally, while this paper examines the usage of seven features, this collection is not by any means enough to cover the complete range of linguistic phenomena in AAVE as a rich dialectal tradition. Additionally, as mentioned above, the overlap between AAVE and other regional and cultural dialects makes it challenging to isolate a model's representation of any given dialect from its representations of other dialects.

\subsection{Future Work}
Future research could focus on generalizing this method of quantifying LLMs' language usage to other dialects, particularly other minoritized dialects. Additionally, future work could involve incorporating a wider range of grammatical features \cite{green2002african}. Finally, most critically, future work could incorporate the discussions expressed in this paper to develop more large-scale, inclusively trained models that can more fluently and accurately depict AAVE, a direction which is beginning to be explored. Enabling models to accurately capture linguistic features of the types we have analyzed is crucial to building more inclusive and linguistically equitable language models. A model's ability to replicate is likely related to its ability to `understand' \cite{devlin2019bert,Radford2019LanguageMA,openai2023gpt4,touvron2023llama,anthropic2024claude}.  If LLMs misrepresent AAVE, then voice assistants, chatbots, and co-writing tools risk alienating Black users, reinforcing inequities in access to AI-driven technologies.

Our prompts explicitly asked for AAVE, possibly affecting how LLMs would use dialectal features in unconstrained generation. Future work could examine spontaneous dialectal usage to test ecological validity, though previous studies have shown model resistance to naturally occurring non-standard dialects \cite{Fleisig:24}.

Future work could also expand the methodology to include not only more features but also phonological and discourse-level features that can capture a more representative picture of AAVE as a dialect.

\section{Conclusion}
We have found that Gemma 2, GPT-4o mini, and Llama 3 can use key features of African American Vernacular English (AAVE), but they struggle to fully capture the nuances of how frequently these features are used and the contexts in which they are used. One possible avenue for improving models along this dimension is augmenting training data with primary sources of minoritized varieties of English, such as AAVE, to better illustrate the grammatical structure of these dialects to the models. 
Improving models' representations of dialectal features is essential for promoting equitable representation of AAVE and other minoritized dialects in language technology and ensuring that linguistic diversity is respected and understood as the field advances. By making dialectal misrepresentation visible and measurable, we hope to guide the development of AI systems that respect and preserve linguistic diversity rather than caricature it.

\subsection{Ethics Statement}

An important question is whether it is ethical to develop models that reflect African American speech. There has been growing concern over LLMs' tendency to build from or even plagiarize the work of human artists, including African American artists, without credit or payment, bringing in discussion of potential language appropriation and commodification \cite{Heller2010commodification,small_black_2023}. Thus, work that aims to improve these models' ability to replicate Black speech should be carefully questioned and scrutinized to ensure that this work does not harm native AAVE speakers nor distort the true origins, intentions, and expression of AAVE. To help think through these concerns, it is important to consider how LLMs already perform at using AAVE. Our current work does not by any means answer questions about whether the field should develop systems that use AAVE or accurately replicate AAVE to human-levels, but we hope that our results can provide empirical data that can help to inform these conversations.

\begin{credits}
\subsubsection{\ackname} This study was funded through the generous support of the Yale Mellon Mays/Edward A. Bouchet Fellowship (YMMB). Special thanks to the first author's YMMB cohort for their support throughout this process.

\subsubsection{\discintname}
The authors declare no competing interests.
\end{credits}
\bibliographystyle{splncs04}

\end{document}